\documentclass[10pt,twocolumn,letterpaper]{article}

\usepackage[pagenumbers]{cvpr} 

\definecolor{cvprblue}{rgb}{0.21,0.49,0.74}
\usepackage[pagebackref,breaklinks,colorlinks,allcolors=cvprblue]{hyperref}

\def\paperID{*****} 
\def\confName{CVPR}
\def\confYear{2026}

\title{
LoR-LUT: Learning Compact 3D Lookup Tables via Low-Rank Residuals
}

\author{Ziqi Zhao \qquad Abhijit Mishra \qquad Shounak Roychowdhury\\
The University of Texas at Austin\\
Austin, TX, USA\\
{\tt\small \{ziqizhao, abhijitmishra\}@utexas.edu, shounak.roychowdhury@ischool.utexas.edu}
}

\begin{document}
\maketitle

\begin{abstract}
We present LoR-LUT, a unified low-rank formulation for compact and interpretable 3D lookup table (LUT) generation. 
Unlike conventional 3D-LUT-based techniques that rely on fusion of basis LUTs, which are usually dense tensors, our unified approach extends the current framework by jointly using residual corrections, which are in fact low-rank tensors, together with a set of basis LUTs.
The approach described here improves the existing perceptual quality of an image, which is primarily due to the technique's novel use of residual corrections. At the same time, we achieve the same level of trilinear interpolation complexity, using a significantly smaller number of network, residual corrections, and LUT parameters.
The experimental results obtained from LoR-LUT, which is trained on the MIT-Adobe FiveK dataset, reproduce expert-level retouching characteristics with high perceptual fidelity and a sub-megabyte model size. 
Furthermore, we introduce an interactive visualization tool, termed LoR-LUT Viewer, which transforms an input image into the LUT-adjusted output image, via a number of slidebars that control different parameters. The tool provides an effective way to enhance interpretability and user confidence in the visual results. 
Overall, our proposed formulation offers a compact, interpretable, and efficient direction for future LUT-based image enhancement and style transfer.

\end{abstract}

\section{Introduction}
\label{sec:intro}

\begin{figure}[!t]
  \centering
  \includegraphics[width=\linewidth]{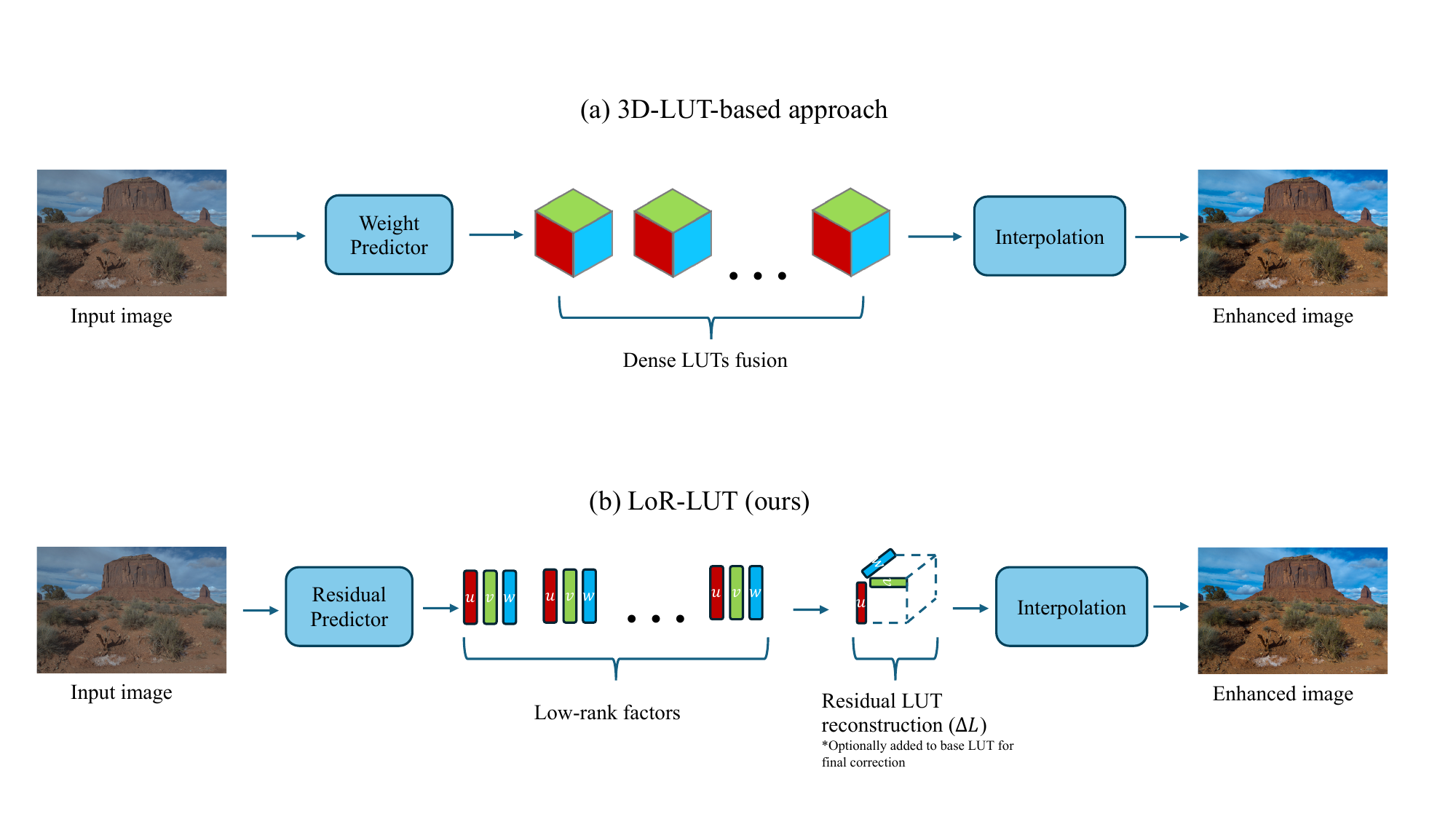}
  \caption{\textbf{Comparison between conventional adaptive 3D-LUTs and our LoR-LUT.}
  (a) conventional adaptive 3D-LUTs rely on fusing several dense basis LUTs.
  (b) LoR-LUT replaces dense fusion with a compact low-rank residual,
  achieving similar flexibility with fewer parameters and identical interpolation complexity.}
  \label{fig:concept}
\end{figure}

Three-dimensional lookup tables (3D LUTs) have been playing an important role in post-production of images and videos because they provide \emph{deterministic, low-latency} color transformations that can be easily embedded in cameras and mobile devices.
A 3D LUT is a sampled representation of the RGB cube, which is a smaller 3D grid of RGB colors, and applies per-pixel interpolation to map input images to desired output images. In this paper we propose a simpler LUT pipeline that is easier to interpret and deploy.

Recent learning-based approaches make LUTs more
image-adaptive -- for instance a CNN-based neural network predicts how to fuse a set of basis LUTs into a single transform (image enhancement) conditioned on image content, achieving expert-level photo enhancement with 4K throughput in the millisecond range~\cite{zeng2022ia3dlut}.
Despite these advantages, two limitations persist. 
\begin{enumerate}
    \item Mainstream image-adaptive 3D LUT methods achieve flexibility primarily via \emph{dense basis LUT fusion}. While dense basis are effective, they are also parameter-heavy and require more storage. It can be redundant given the \emph{low-dimensional structure} that exists in many images (exposure, contrast, color balance).
    \item Capturing \emph{local, spatially varying} adjustments that are important for scenes with mixed illumination or distinct semantic regions that usually requires additional spatial modules (e.g., spatial-aware predictors or bilateral grids), which inflate learning parameters that substantially impact computational runtime.~\cite{wang2021sa3dlut,kim2024bilateral}.
\end{enumerate}

Currently this area is getting a lot of attention as
several researchers are trying to solve these issues. \emph{Sampling-aware} designs learn non-uniform grid intervals to allocate samples where the transform is highly nonlinear~\cite{yang2022adaint}. \emph{Factorized/separable} designs decompose color transforms into efficient 1D/3D components to reduce memory and training difficulty~\cite{yang2022seplut}. \emph{Spatial-aware} variants inject local information using lightweight two-head predictors or \emph{bilateral grids} that decouple spatial reasoning from color mapping~\cite{wang2021sa3dlut,kim2024bilateral}. Meanwhile, \emph{implicit} formulations replace explicit grids with continuous neural functions, improving memory efficiency and style blending but sacrificing direct cube export and some deployment benefits~\cite{conde2024nilut}.

We aim to keep what makes explicit LUTs compelling---\emph{speed, exportability, interpretability}---while replacing dense, potentially redundant parameters with a \emph{compact, structured} alternative. As illustrated in Figure ~\ref{fig:concept}, LoR-LUT replaces the dense fusion
of multiple basis LUTs with a compact low-rank residual, maintaining the same
interpolation complexity while reducing parameters.
We introduce \textbf{LoR-LUT}, a \emph{unified low-rank formulation} for learning \emph{compact and interpretable} 3D LUTs. Rather than relying solely on dense basis fusion, LoR-LUT augments (or even replaces) bases with a \emph{low-rank residual} generated per image by a tiny hyper-network. Concretely, we parameterize the residual as a sum of a handful of \emph{rank-1 components} (a CP-style decomposition along the three LUT axes), which directly updates the explicit 3D grid before the standard interpolation operator is applied. This preserves the classic \emph{per-pixel interpolation complexity} (e.g., trilinear) while substantially reducing learnable parameters and improving editability of the transform. In practice, we find that the low-rank residual \emph{captures the ``last-mile'' nonlinearity} that dense bases often redundantly encode.

Beyond parsimony, the low-rank residual offers two further advantages. \textbf{(i)} It is \emph{naturally interpretable}: each rank-1 component acts like a separable ``brush'' in RGB space, and the sum of a few such components yields readable patterns on LUT slices, which we expose in our \emph{LoR-LUT Viewer} for interactive inspection and control. \textbf{(ii)} The residual-only variant (\emph{K=0}) is surprisingly strong in our experiments, suggesting that much of the enhancement transform lies in a \emph{low-dimensional manifold} that does not require dense bases at all, opening a new compact regime for image-adaptive LUTs while preserving deployment properties.

Our technique, LoR-LUT, reproduces expert-style edits with \emph{high perceptual fidelity}. We validate our technique of a smaller model size using \emph{MIT-Adobe FiveK}~\cite{bychkovsky2011fivek} dataset, which is the de-facto benchmark for expert retouching. The FiveK dataset enables fair comparison to (1) image-adaptive fusion, (2) sampling-aware (AdaInt), (3) separable (SepLUT), (4) spatial-aware (SA-3D-LUT), and other baselines (HDRNet, CSRNet, etc.).
We have two interpolation choices: trilinear \cite{zeng2022ia3dlut} or tetrahedral. We use trilinear interpolation for easier computation compared to tetrahedral interpolation.

\noindent The main contributions of this paper are:
\begin{itemize}
    \item \textbf{Unified low-rank residual formulation:} We propose LoR-LUT, a compact, image-adaptive 3D LUT that augments or replaces dense bases with a \emph{learned low-rank residual}. The residual is predicted per image by a small network, and it is applied once to the sampled 3D grid before interpolation, thereby \emph{preserving the classic LUT pipeline} and deployment friendliness. A residual-only LoR-LUT attains strong results, evidencing a \emph{low-dimensional structure} in expert retouching transforms even without basis LUTs.
    \item \textbf{Compactness without sacrificing speed/portability:} By shifting capacity into a small number of rank-1 components, LoR-LUT reduces parameters while keeping the \emph{same interpolation complexity} as conventional explicit LUTs.
    

    \item \textbf{Tool development:} We showcase an interactive tool, termed \emph{LoR-LUT Viewer}, that displays the live residual structures (rank, magnitude, and components).
    \item \textbf{Comprehensive evaluation:} On FiveK, LoR-LUT achieves expert-level perceptual quality with a sub-MB model, and compares favorably to representative families: image-adaptive fusion~\cite{zeng2022ia3dlut}, sampling-aware~\cite{yang2022adaint}, separable~\cite{yang2022seplut}, and spatial-aware~\cite{kim2024bilateral}.
\end{itemize}

\section{Method}
\label{sec:method}

In this section, we propose our unified \textbf{LoR-LUT} formulation which integrates a compact low-rank residual representation into the standard LUT pipeline.
In the following subsections, we discuss starting from the fundamentals of 3D Lookup Tables, low-rank representations, and explore unified LoR-LUT network architecture, training objectives, and computational characteristics in detail.

\subsection{3D LUT and Trilinear Interpolation}
\label{subsec:trilinear}

\begin{figure}[t]
  \centering
  \includegraphics[width=\linewidth]{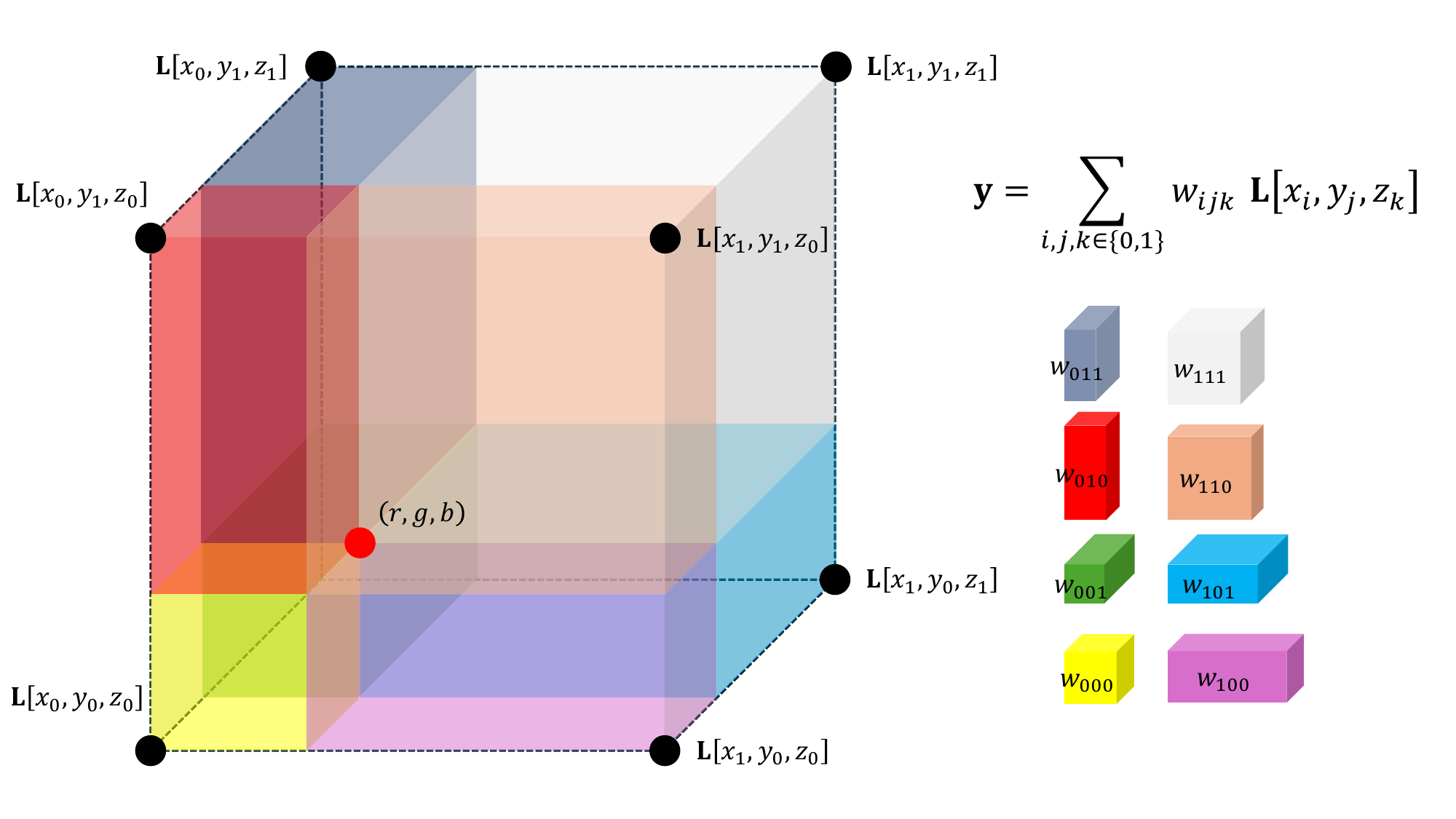}
  
  \caption{
Trilinear interpolation in a 3D LUT.
Each input color $(r,g,b)$ lies within a cube defined by eight lattice vertices $\mathbf{L}[x_i,y_j,z_k]$.
The output color $\mathbf{y}$ is a weighted sum of eight vertices of the cube,
where $w_{ijk}$ is a trilinear interpolation weight that decreases with distance from $(r,g,b)$ along each axis.
}
\label{fig:trilinear}
\end{figure}

As illustrated in Figure~\ref{fig:trilinear}, 
a 3D LUT maps an input color $\mathbf{x}=(r,g,b)$ in normalized RGB space to an output color $\mathbf{y}=(r',g',b')$ 
through sampling on a discrete cube of grid size $G$.
Each lattice point stores a color vector $\mathbf{L}[i,j,k]\in[0,1]^3$.
Given an arbitrary input $\mathbf{x}$, the output color is computed by interpolating the eight surrounding lattice vertices:
\begin{equation}
\mathbf{y} = 
\sum_{i=0}^1\sum_{j=0}^1\sum_{k=0}^1 
w_{ijk} \, \mathbf{L}[x_i, y_j, z_k],
\label{eq:trilinear}
\end{equation}
where $w_{ijk}$ are the trilinear interpolation weights determined by the fractional distances of $\mathbf{x}$ to its nearest grid coordinates.
This operation is differentiable and hardware-friendly, enabling fast inference on both CPUs and GPUs.

Despite its simplicity and computational speed, the expressive capacity of a single LUT is limited by its grid resolution and discrete sampling.
To enhance representation power, recent methods learn multiple basis LUTs and use a small CNN to predict their fusion weights for each image~\cite{zeng2022ia3dlut}.
However, fusing several dense LUTs greatly increases parameters and may redundantly capture correlated color variations.

\subsection{Low-Rank Residual Representation}
\label{subsec:lowrank}

\begin{figure}[t]
  \centering
  \includegraphics[width=\linewidth]{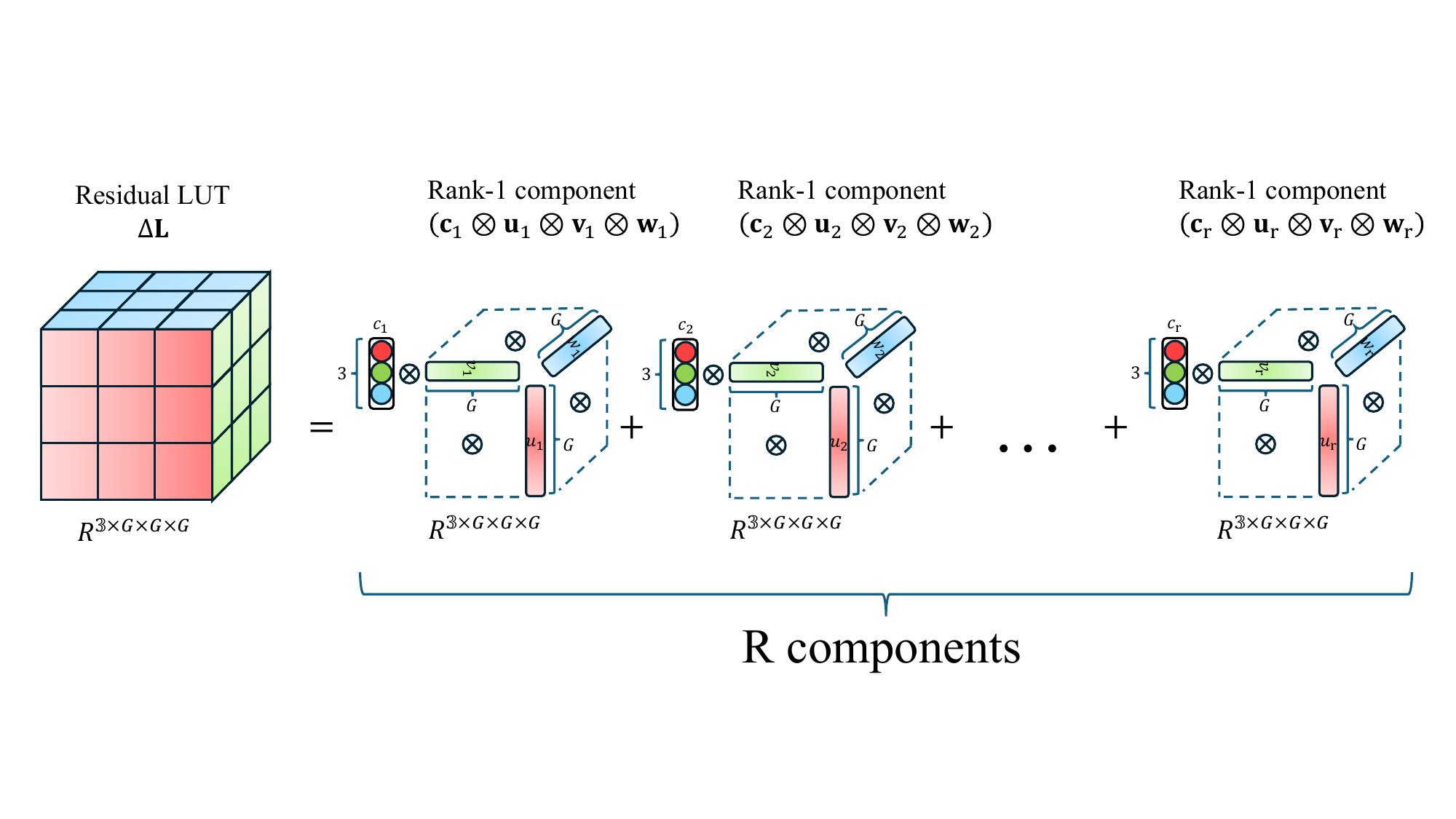}
  
  \caption{\textbf{Low-rank residual representation.}
  Each rank-1 component factorizes the correction tensor into three axis-specific vectors and a color coefficient. 
  Summing $R$ such components yields a compact yet expressive residual tensor $\Delta\mathbf{L}$.}
  \label{fig:lowrank}
\end{figure}

We present a low-rank residual representation using Canonical Polyadic Decomposition (CPD) \cite{kolda2009tensor} that augments the standard 3D LUT with a \emph{low-rank residual correction}:
\begin{equation}
\Delta \mathbf{L} = 
\sum_{r=1}^{R} \mathbf{c}_r \otimes 
\mathbf{u}_r \otimes 
\mathbf{v}_r \otimes 
\mathbf{w}_r,
\label{eq:cp}
\end{equation}
where $G$ is the number of sample points on an axis of a LUT, and $\mathbf{u}_r,\mathbf{v}_r,\mathbf{w}_r\!\in\!\mathbb{R}^G$ are axis factors, $\mathbf{c}_r\!\in\!\mathbb{R}^3$ is the color coefficient, and $\otimes$ denotes the outer product.
The rank $R$ controls model capacity.
The structure of this low-rank factorization is visualized in Figure~\ref{fig:lowrank}.
Instead of fusing multiple dense basis LUTs, we generate a lightweight correction tensor $\Delta \mathbf{L}$ to refine a base LUT $\mathbf{L}_0$:
\begin{equation}
\mathbf{L}^* = \mathbf{L}_0 + \Delta \mathbf{L}.
\end{equation}

This low-rank design yields several benefits:
(1) drastically fewer parameters ($3G R + 3R$ vs.\ $3G^3$ for dense LUTs);
(2) continuous controllability through $R$; and
(3) interpretability—each rank-1 component corresponds to a separable ``color brush'' in the RGB cube.
During inference, the residual tensor is reconstructed on the fly and added to the base LUT, followed by the same trilinear interpolation \eqref{eq:trilinear}.
Thus, LoR-LUT preserves the original interpolation complexity while improving expressiveness.

\subsection{Unified LoR-LUT Architecture}
\label{subsec:architecture}

\begin{figure*}[t]
  \centering
  \includegraphics[width=\textwidth]{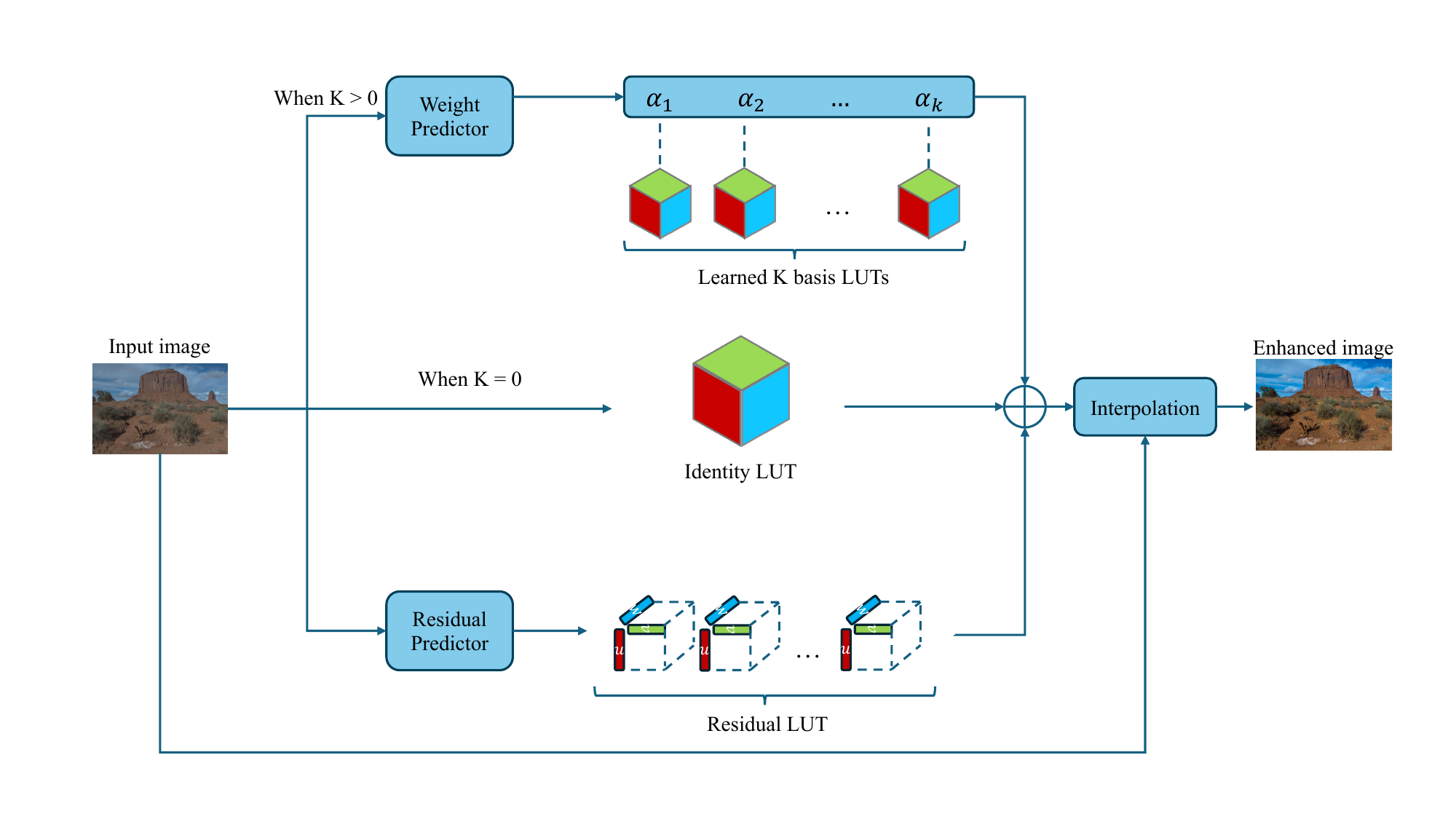}
  
  \caption{\textbf{Overall architecture of LoR-LUT.}
  Two lightweight networks predict the fusion weights and low-rank residual factors from the input image.
  The reconstructed LUT is then applied using the standard trilinear interpolation operator.
  The entire pipeline is differentiable and end-to-end trainable.}
  \label{fig:architecture}
\end{figure*}

The complete LoR-LUT system is illustrated in Figure~\ref{fig:architecture}.
Given an input image $\mathbf{I}$, two lightweight predictors are employed:

\begin{itemize}
  \item \textbf{Weight predictor} $f_\alpha(\mathbf{I})$: a shallow CNN that estimates the fusion weights $\boldsymbol{\alpha}\in\mathbb{R}^K$ for $K$ base LUTs.
  \item \textbf{Residual predictor} $f_\text{res}(\mathbf{I})$: an MLP that outputs the factor vectors $\{\mathbf{u}_r,\mathbf{v}_r,\mathbf{w}_r,\mathbf{c}_r\}_{r=1}^R$ for the low-rank residual.
\end{itemize}

The final LUT is obtained as:
\begin{equation}
\mathbf{L}^*(\mathbf{I}) 
= \sum_{k=1}^K \alpha_k(\mathbf{I})\, \mathbf{L}_k
+ \Delta\mathbf{L}(\mathbf{I}),
\end{equation}
and applied to the input via trilinear interpolation to yield the enhanced output $\hat{\mathbf{I}}$.

\begin{table*}[t]
\centering
\caption{\textbf{Network architecture details.} 
Layers used in the weight and residual predictors of LoR-LUT, 
together with the learnable basis LUTs when $K>0$. 
Here $G$ is the LUT grid size and $R$ is the low-rank decomposition rank.}
\label{tab:architecture}
\vspace{2mm}
{\small
\setlength{\tabcolsep}{4pt}
\begin{tabular}{lcc}
\toprule
\textbf{Module} & \textbf{Configuration} & \textbf{Output} \\
\midrule

Weight predictor 
& Conv($3\!\rightarrow\!16$), Conv($16\!\rightarrow\!32$), AvgPool, FC($32\!\rightarrow\!K$)
& $\alpha \in \mathbb{R}^{K}$ \\

Residual predictor 
& Conv($3\!\rightarrow\!16$), Conv($16\!\rightarrow\!32$), AvgPool, FC$\times4$ 
& $R\times(3G + 3)$ \\

Basis LUTs (if $K>0$)
& $K$ learnable tensors of size $G^3\times3$
& $K\times 3G^{3}$ \\

\bottomrule
\end{tabular}
}
\end{table*}

The total number of learnable parameters ($\#\theta_{\text{total}}$) in LoR-LUT consists of
(1) the weight predictor ($\#\theta_{\text{weight}}$),
(2) the residual predictor ($\#\theta_{\text{residual}}$),
and (3) the optional learnable basis LUTs when $K>0$ ($\#\theta_{\text{LUT}}$).

\textbf{Weight predictor:}
The predictor ends with a fully-connected layer
$\mathrm{FC}(32 \!\rightarrow\! K)$
that maps a 32-dimensional global feature vector to $K$ fusion weights
$\alpha \in \mathbb{R}^{K}$.
A linear layer with input dimension 32 and output dimension $K$ contains
$32K$ weights and $K$ biases, giving
\[
\#\theta_{\text{weight}} 
= 5088 + (32K + K)
= 5088 + 33K,
\]
where $5088$ accounts for the two convolutional layers in the predictor.
The $33K$ term reflects that each of the $K$ outputs requires
32 weights and one bias; these are \emph{prediction parameters}
rather than the fusion weights themselves.

\textbf{Residual predictor:}
Similarly, the residual predictor produces the low-rank factors
$\{\mathbf{u}_r, \mathbf{v}_r, \mathbf{w}_r, \mathbf{c}_r\}_{r=1}^R$.
For grid size $G$, the three factor heads
$\mathrm{FC}(32 \!\rightarrow\! RG)$
and the color head
$\mathrm{FC}(32 \!\rightarrow\! 3R)$
contain
\[
3(33RG) + 99R
= 99R(G+1)
\]
parameters, in addition to the $5088$ parameters in the convolutional
encoder.  
Thus,
\[
\#\theta_{\text{residual}}
= 5088 + 99R(G+1).
\]

\textbf{Learnable basis LUTs:}
When $K>0$, the model includes $K$ learnable 3D LUTs, each of size
$G^3\times 3$, contributing
\[
\#\theta_{\text{LUT}} = 3K G^{3}
\]
parameters.

\textbf{Total:}
Combining the above,
the full model contains
\[
\#\theta_{\text{total}}
= 10176 + 99R(G+1) + K\,(33 + 3G^{3})
\]
For the commonly used setting $G{=}33$,
this simplifies to
\[
\#\theta_{\text{total}}
= 10176 + 3366R + 107{,}844K.
\]

\noindent
Table~\ref{tab:architecture} summarizes the network configuration used in all experiments. 
The weight and residual predictors are extremely lightweight---together they contain 
only $5088{+}33K + 5088{+}R(G+1)$ parameters 
(5088 is a fixed number of learnable parameters used inside CNN)(e.g., $\sim\!37$K for $K{=}0,R{=}8$ or $\sim\!118$K for $K{=}0,R{=}32$). When $K{>}0$, the model includes $K$ learnable 3D LUTs, 
each with $3G^3$ parameters ($107{,}811$ for $G{=}33$), 
which dominate the total parameter count. 
Both predictors operate on a low-resolution input (e.g., $512{\times}512$) 
and incur negligible runtime compared to the 3D-LUT lookup; 
thus the overall latency is primarily determined by the LUT application 
evaluated in Section~\ref{subsec:efficiency}.

\subsection{Training Objectives}
\label{subsec:loss}

We train LoR-LUT in a supervised manner on paired datasets such as MIT-Adobe FiveK~\cite{bychkovsky2011fivek}.
Given a ground-truth enhanced image $\mathbf{I}_{\text{gt}}$, the total loss is:
\begin{equation}
\begin{aligned}
\mathcal{L} 
&= \lambda_1 \, \|\hat{\mathbf{I}} - \mathbf{I}_{\text{gt}}\|_1
   + \lambda_2 \, \mathcal{L}_{\text{LPIPS}}
   + \lambda_3 \, \mathcal{L}_{\Delta E_{00}} \\
&\quad + \lambda_4 \, \mathcal{L}_{\text{TV}}(\mathbf{L}^*)
   + \lambda_5 \, \|\Delta\mathbf{L}\|_2^2 .
\end{aligned}
\label{eq:loss}
\end{equation}
where $\mathcal{L}_{\text{LPIPS}}$~\cite{zhang2018lpips} encourages perceptual similarity,
$\mathcal{L}_{\Delta E_{00}}$~\cite{sharma2005ciede2000} ensures accurate color reproduction,
$\mathcal{L}_{\text{TV}}$ regularizes LUT smoothness to avoid quantization artifacts,
and the $\ell_2$ penalty constrains the residual magnitude for stability.

\begin{figure}[t]
  \centering
  \includegraphics[width=\linewidth]{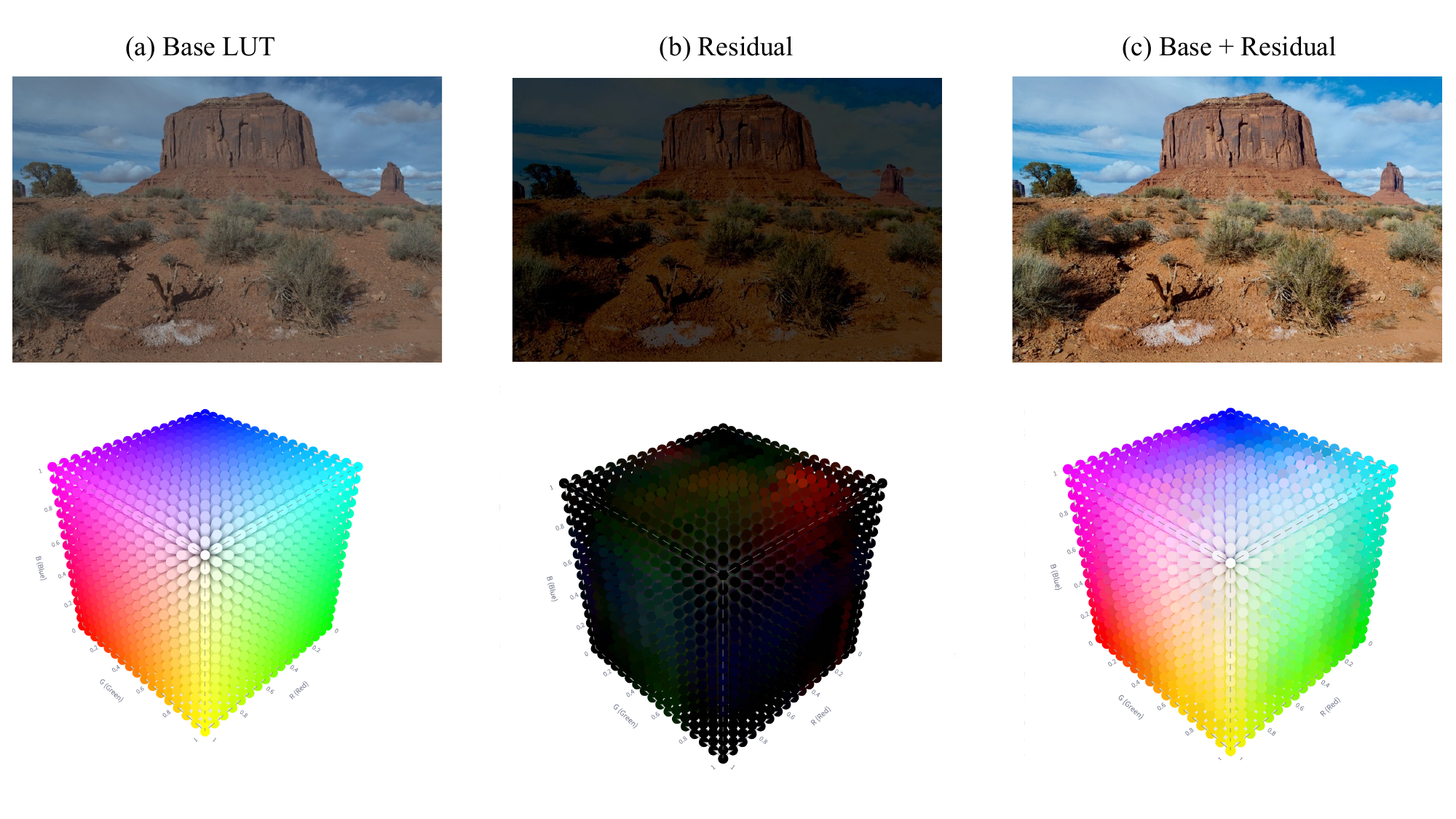}
  \caption{\textbf{Visualization of low-rank residual correction in LoR-LUT.}
  (a) Base LUT (identity, $K{=}0$); (b) residual output; and (c) combined base+residual result.
  The bottom row shows the corresponding 3D LUT cubes. 
  The low-rank residual introduces subtle yet structured color shifts.}
  \label{fig:residual_vis}
\end{figure}

The above loss terms jointly enforce accurate, perceptually pleasing, and stable color mappings.
In practice, we set $(\lambda_1,\lambda_2,\lambda_3,\lambda_4,\lambda_5)=(1,0.0,0.0,0.001,0.001)$.
Training converges within 100k iterations using AdamW optimizer with cosine learning rate decay.

\subsection{Complexity Analysis}
\label{subsec:complexity}


\begin{table}[t]
  \centering
  \caption{\textbf{Model size comparison.}
  Number of learnable parameters for representative image-adaptive 3D-LUT methods.
  All methods use a LUT grid size of $G{=}33$.}
  \label{tab:complexity}
  \vspace{2mm}
  \setlength{\tabcolsep}{5pt}
  \begin{tabular}{lcc}
  \toprule
  \textbf{Method} & \textbf{Params} & \textbf{Interp.} \\
  \midrule
  IA-3D-LUT~\cite{zeng2022ia3dlut} & 0.59M & Trilinear \\
  AdaInt~\cite{yang2022adaint}    & 0.61M & Adaptive \\
  SA-3D-LUT~\cite{wang2021sa3dlut}     &4.5M & Trilinear\\
  \textbf{LoR-LUT (ours)}         & \textbf{37K} & Trilinear \\
  \bottomrule
  \end{tabular}
\end{table}

LoR-LUT inherits the same interpolation complexity $\mathcal{O}(1)$ per pixel as a standard 3D-LUT.
The additional cost stems only from reconstructing the low-rank residual tensor, which involves $R$ outer products of size $G$ per axis.
As summarized in Table~\ref{tab:complexity}, LoR-LUT uses fewer parameters than existing adaptive 3D-LUT variants (IA-3DLUT, AdaInt) while preserving the same explicit trilinear LUT pipeline.

\section{Experiments}
\label{sec:experiments}

We present our experiments that shows the efficacy and interpretability of LoR-LUT.
All results are produced using our public implementation.\footnote{Code and pretrained models will be released in the final version of the paper}.


\subsection{Datasets and Implementation}
\label{subsec:datasets}

We train and evaluate LoR-LUT primarily on the MIT-Adobe FiveK dataset~\cite{bychkovsky2011fivek}, 
which contains 5{,}000 RAW photographs retouched by five experts.
Following prior works~\cite{zeng2022ia3dlut,yang2022adaint,yang2022seplut}, 
we use expert~C's rendition as the ground truth and resize images to $480$p for training.

We implement the LoR-LUT Viewer (Figure~\ref{fig:tool}). In the viewer, the LUT grid resolution is set to $G{=}33$, because it is a commonly used user-defined value.
For the low-rank residual, we set rank $R{=}8$ unless otherwise noted.
Both the weight and residual predictors are lightweight CNNs with fewer than 0.5M parameters.
Training is performed using AdamW optimizer for 100k iterations, 
with a batch size of~4, initial learning rate $0.0001$, and cosine decay.
We use the loss terms described in Eq.~(\ref{eq:loss}) with 
$(\lambda_1,\lambda_2,\lambda_3,\lambda_4,\lambda_5)=(1,0.0,0.0,0.001,0.001)$.

\paragraph{Evaluation metrics.}
We report PSNR, SSIM, LPIPS~\cite{zhang2018lpips}, 
and $\Delta E_{00}$~\cite{sharma2005ciede2000}, 
following standard practice in photo retouching literature.

\subsection{Quantitative Comparison}
\label{subsec:quantitative}

We compare LoR-LUT with representative LUT-based and non-LUT baselines:
IA-3DLUT~\cite{zeng2022ia3dlut}, 
AdaInt~\cite{yang2022adaint}, 
SepLUT~\cite{yang2022seplut}, 
HDRNet~\cite{gharbi2017hdrnet}, 
and DPE~\cite{chen2018dpe}.
All methods are retrained on FiveK using official code or configurations.


\begin{table}[t]
  \centering
  \caption{\textbf{Quantitative comparison on MIT-Adobe FiveK (Expert C).}
  LoR-LUT achieves the best perceptual quality}
  \label{tab:benchmark}
  \vspace{1mm}
  {\small
  \begin{tabular}{lccccc}
  \toprule
  \textbf{Method} & PSNR$\uparrow$ & SSIM$\uparrow$ & LPIPS$\downarrow$ \\
  \midrule
  HDRNet~\cite{gharbi2017hdrnet} & 22.15 & 0.840 & 0.182 \\
  DPE~\cite{chen2018dpe} & 23.75 & 0.908 & 0.094  \\
  IA-3D-LUT~\cite{zeng2022ia3dlut} & 22.27 & 0.837 & 0.183 \\
  AdaInt~\cite{yang2022adaint} & 25.13 & 0.921 & - \\
  SepLUT~\cite{yang2022seplut} & 25.32 & 0.918 & - \\
  \textbf{LoR-LUT} \\[-3pt] {\scriptsize (K=0,R=32)} & \textbf{25.53} & \textbf{0.901} & \textbf{0.083}  \\
  \textbf{LoR-LUT} \\[-3pt] {\scriptsize (K=0, R=8)} & 25.35 & 0.901 & 0.079  \\
  \bottomrule
  \end{tabular}
  }
\end{table}

As shown in Table~\ref{tab:benchmark}, 
LoR-LUT surpasses previous LUT-based approaches in both PSNR and perceptual metrics while 
using fewer parameters.
Notably, the residual-only configuration ($K{=}0$) still performs competitively,
revealing that much of the color transformation can be captured in a low-dimensional subspace.

\subsection{Qualitative Comparison}
\label{subsec:qualitative}

\begin{figure*}[t]
  \centering
  \includegraphics[width=\textwidth]{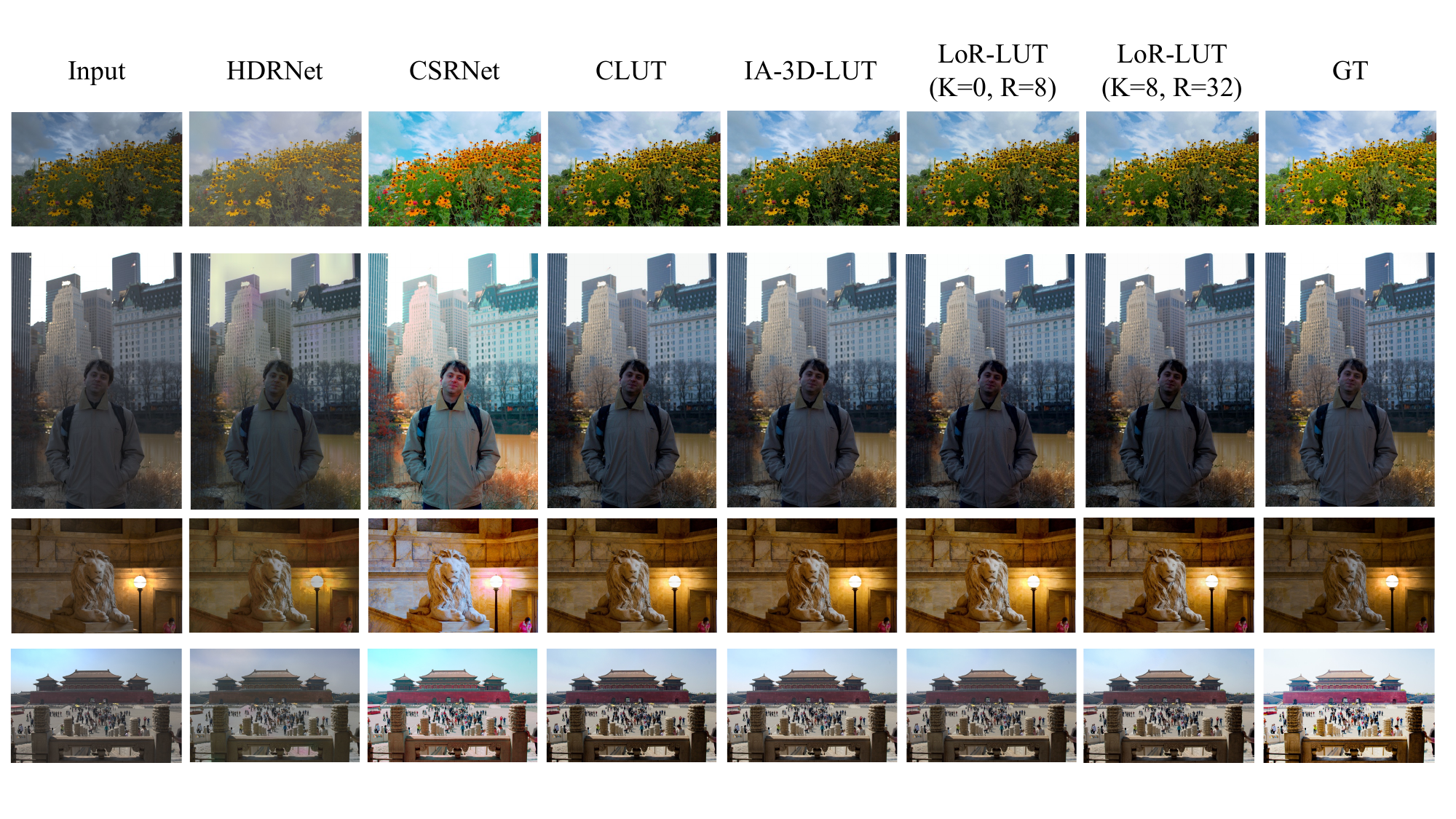}
  
  \caption{
\textbf{Qualitative comparison on MIT-Adobe FiveK (Expert C).}
From left to right: input image, HDRNet~\cite{gharbi2017hdrnet}, 
CSRNet~\cite{CSRNet}, CLUT~\cite{zhang2022clutnet}, IA-3DLUT~\cite{zeng2022ia3dlut}, 
our residual-only variant (LoR-LUT, $K{=}0$), our full model (LoR-LUT), 
and the expert ground truth (GT).
Compared to prior LUT-based and CNN-based baselines, LoR-LUT 
produces more natural color tones, smoother local contrast, and 
better overall perceptual fidelity, closely matching the expert edit.
}
\label{fig:qualitative}
\end{figure*}

Figure~\ref{fig:qualitative} illustrates visual comparisons on representative FiveK examples.
Our LoR-LUT reproduces expert retouching with natural contrast and faithful color tones.
Compared with IA-3DLUT or AdaInt, it yields smoother color gradients and less over-saturation.
The residual-only LoR-LUT (K=0) maintains global fidelity but slightly under-enhances fine contrast, confirming the complementary role of the base LUTs.

\subsection{Ablation Studies}
\label{subsec:ablation}

Our ablation studies analyze the effect of the low-rank residual and the number of basis LUTs. Table~\ref{tab:rank} reports the performance of residual-only LoR-LUT ($K=0$) with different ranks $R$. Increasing $R$ from 4 to 8 and 32 consistently improves PSNR and $\Delta E_{00}$ while keeping the parameter count below 0.12M. The $R=8$ model already reaches 25.35 dB with a very compact 0.04M parameters. These results support our claim that most photographic color transformations reside in a low-dimensional manifold that can be well captured by a small number of low-rank components.



\begin{table}[t]
\centering
\caption{Effect of the low-rank $R$ for the residual-only LoR-LUT ($K\!=\!0$). }
\label{tab:rank}
\begin{tabular}{c c c c c c}
\toprule
$R$ & PSNR$\uparrow$ & SSIM$\uparrow$ & LPIPS$\downarrow$ & $\Delta E_{00}\downarrow$ & Params \\
\midrule
4   & 25.13 & 0.895 & 0.081 & 6.16 & 0.024M \\
8   & 25.35 & 0.901 & \textbf{0.079} & 5.85 & 0.037M \\
32  & \textbf{25.53} & \textbf{0.901} & 0.083 & \textbf{5.75} & 0.118M \\
\bottomrule
\end{tabular}
\end{table}



\subsection{Efficiency and Parameter Analysis}
\label{subsec:efficiency}

We benchmark runtime and model size under 4K resolution. As summarized in Table \ref{tab:complexity2}, LoR-LUT keeps the model compact (only 0.12M parameters for the residual-only variant) while preserving fast LUT-style inference: on an NVIDIA T4 GPU, a 4K image (3840×2160) is processed in about 68 ms, and a 1080p frame in 17.8 ms. The low-rank residual reconstruction adds less than 1 ms on top of the LUT lookup, and different choices of $(K,R)$ barely affect the overall latency, confirming that the cost is dominated by the 3D-LUT application.


\begin{table}[t]
  \centering
  \caption{Runtime and parameter analysis of LoR-LUT variants on 4K inputs
  (3840$\times$2160). Latency is averaged over 100 runs on NVIDIA T4 and L4
  GPUs. All models use grid size $G{=}33$.}
  \label{tab:complexity2}
  \begin{tabular}{lcccc}
    \toprule
    Method & Params & Model size & 4K latency (T4)\\
    \midrule
    LoR-LUT \\[-3pt] {\scriptsize (K=0,R=32)} & 0.12M & 0.45MB & 68.3ms\\
    LoR-LUT \\[-3pt] {\scriptsize (K=8, R=32)} & 0.98M & 3.74MB & 69.0ms\\
    LoR-LUT \\[-3pt] {\scriptsize (K=8, R=0)}  & 0.87M & 3.33MB & 68.1ms \\
    \bottomrule
  \end{tabular}
\end{table}

\subsection{Visualization and Interpretability}
\label{subsec:visualization}
To further 
analyze what the network has learned, figure~\ref{fig:residual_vis} decomposes the residual into 
R rank-1 components. For each component we visualize its three 1D factor 
curves along the R/G/B axes and its RGB coefficient vector. These patterns are exposed to users through the interactive LoR-LUT Viewer (Figure~\ref{fig:tool}). In practice, these 
components correspond to intuitive photographic operations—such as warming 
highlights or slightly cooling shadows—and their sum yields a smooth global 
transform in the LUT cube.

\section{LoR-LUT Viewer}
\begin{figure}[t]
  \centering
  \includegraphics[width=\linewidth]{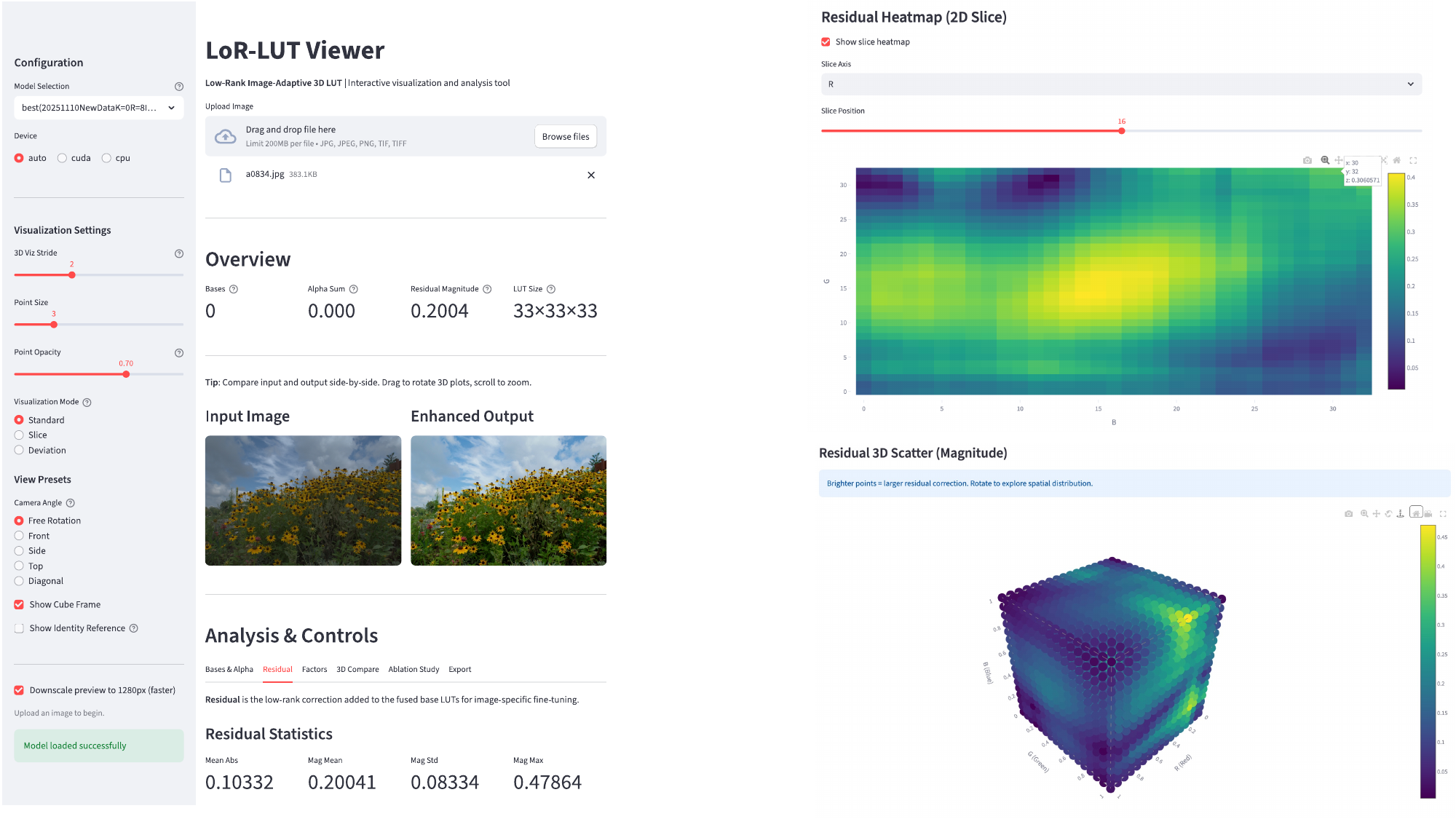}
  
  \caption{\textbf{LoR-LUT Viewer for exploring low-rank residuals.}
  The web-based tool takes an input image, displays the enhanced output and the 
corresponding 3D LUT cube, and exposes sliders for the magnitude of each rank-1 
component.}
  \label{fig:tool}
\end{figure}

In this research, we developed a web-based interactive tool, termed \textbf{interactive LoR-LUT Viewer} (Figure~\ref{fig:tool}), 
which allows users to manipulate the magnitude or orientation of each rank component
and instantly preview the resulting color transformations. Using this tool, we are able to compare and interpret results before and after enhancement, perform ablation studies, and analyze statistical information.
\label{sec:tool}

\begin{figure}[t]
  \centering
  \includegraphics[width=\linewidth]{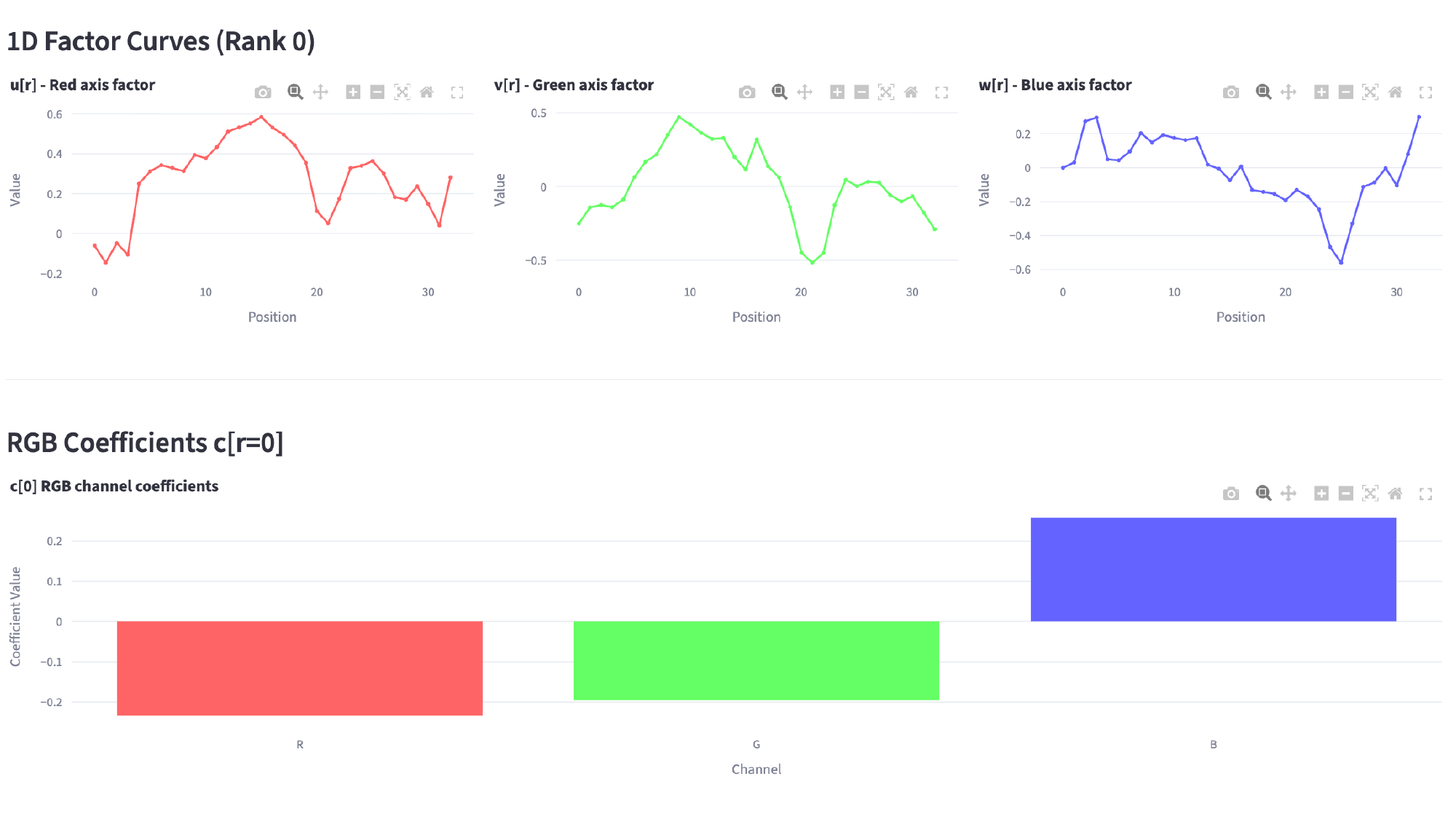}
  
  \caption{\textbf{Low-rank residual factors on an example image.}
  Top: for each rank-1 component, we plot its three 1D factor curves along the 
R, G, and B axes of the LUT grid, revealing separable tonal trends such as 
highlight boosts or shadow roll-off. Bottom: the corresponding RGB 
coefficients indicate how strongly each component acts on each color channel. 
Together, these factors make individual components interpretable as 
"color brushes" in the 3D LUT cube.}
  \label{fig:residual_vis}
\end{figure}

\section{Discussion}
\label{sec:discussion}

The proposed LoR-LUT framework revisits the 3D-LUT pipeline 
from the perspective of low-rank decomposition and residual learning.
Below we discuss few empirical findings and their implications.

A surprising observation from our experiments is that the \textbf{residual-only} configuration ($K{=}0$) already yields strong performance,
sometimes on par with the full model (Table~\ref{tab:benchmark}).
This indicates that a large portion of photographic color transformations
resides in a \emph{low-dimensional subspace} of the 3D color cube.
In other words, the learned rank-$R$ components provide sufficient bases to span 
the manifold of expert-level tonal mappings.
This property aligns with the physical intuition that most global edits—exposure, contrast, and color balance—are separable along RGB axes and can be efficiently modeled by a small number of rank-1 terms.



LoR-LUT maintains the same $\mathcal{O}(1)$ trilinear interpolation cost per pixel as classical LUTs.
The low-rank reconstruction overhead is negligible ($<$0.5\,ms per 4K image). For the residual-only configuration ($K{=}0$), the total model remains below 1 MB,
making it suitable for integration into camera pipelines or mobile ISPs.

Future work may explore integrating bilateral grids~\cite{kim2024bilateral} 
or attention-guided rank modulation to extend LoR-LUT to spatially adaptive enhancement.

\section{Related Work}
Our work is motivated by other works in the area of learnable LUT design.
\begin{itemize}
\item{[Image-adaptive LUTs]}
Learning image-adaptive 3D LUTs (3D-LUTs) has emerged as an efficient alternative to heavy end-to-end CNNs for photo enhancement.
Zeng \etal learn a small set of basis 3D-LUTs plus a lightweight CNN that predicts content-dependent fusion weights to form an image-specific LUT, achieving real-time inference with strong fidelity~\cite{zeng2022ia3dlut}.
Subsequent works improve this paradigm along complementary axes: 
(1) \emph{separable color transforms} that factor component-independent curves and component-correlated interactions (SepLUT) to boost cell utilization with smaller LUTs~\cite{yang2022seplut}; 
(2) \emph{non-uniform sampling} in the color lattice (AdaInt) that allocates more bins to highly non-linear regions while keeping sparse sampling where transforms are near-linear~\cite{yang2022adaint}.
These advances underscore that better parameterization---not necessarily bigger LUTs---drives quality/efficiency.

\item{[Spatially-aware LUTs]}
Pure color-to-color LUTs ignore spatial context, which can limit local tone mapping and edge-aware edits.
Two lines of work inject spatial information while preserving efficiency.
Wang \etal learn \emph{spatial-aware} 3D-LUTs with a two-head predictor that fuses global scenario and pixel-wise category cues for real-time enhancement~\cite{wang2021sa3dlut}.
More recently, Kim and Cho use \emph{bilateral grids} to derive spatial features by non-parametric slicing, then combine them with predicted 3D-LUTs; they further compare interpolation choices and report that tetrahedral works best for 3D-LUT transforms while trilinear suits bilateral slicing~\cite{kim2024bilateral}.
These designs show that spatial priors can be introduced without ballooning model size.

\item{[Implicit and compressed LUT representations]}
Orthogonal to spatial modeling, another thread targets memory/computation.
CLUT-Net compresses 3D-LUTs via learnable low-dimensional representations and transformation matrices, delivering large parameter savings with minimal accuracy loss~\cite{zhang2022clutnet}.
NILUT parameterizes the color transform as a continuous \emph{neural implicit} function conditioned on style, enabling compact storage and smooth style blending while emulating real LUT behavior~\cite{conde2024nilut}.
Complementarily, ICE-LUT “tames” training-time pointwise networks and converts them into \emph{pure LUT} operators at deployment, removing convolutions for extremely fast, hardware-agnostic inference~\cite{yang2024icelut}.
Together, these works explore the spectrum from compressed explicit tables to continuous implicit fields and fully LUT-only inference.

\item{[Classic non-LUT baselines]}
Outside the LUT family, lightweight operators provide strong anchors.
HDRNet predicts locally affine color transforms in a bilateral grid and slices them back to full resolution for mobile real-time enhancement~\cite{gharbi2017hdrnet}.
Unpaired and GAN-based pipelines such as DPE pursue personalized retouching without paired supervision~\cite{chen2018dpe}.
While not LUT-based, these methods are commonly adopted as reference baselines for efficiency and perceptual quality on photo retouching benchmarks.

\item{[Interpolation practice and color science]}
In practical color pipelines, interpolation choice materially affects banding and precision for modest LUT sizes.
Industry and academic reports often recommend \emph{tetrahedral} interpolation for 3D-LUTs due to better numerical behavior than trilinear at the same resolution, a trend also corroborated in recent spatial-LUT studies~\cite{kim2024bilateral,vandenberg2017aces}.
For perceptual evaluation, LPIPS~\cite{zhang2018lpips} complements PSNR/SSIM, and $\Delta E_{00}$~\cite{sharma2005ciede2000} remains a standard color-difference metric that correlates with human judgments.

\item{[Datasets and benchmarks]}
MIT-Adobe FiveK is the de facto supervised benchmark with expert-level retouching pairs~\cite{bychkovsky2011fivek}.
PPR10K targets portrait retouching with large-scale, high-resolution annotations~\cite{liang2021ppr10k}.
Following prior art, we report PSNR/SSIM/LPIPS and $\Delta E_{00}$ on these datasets and compare against representative LUT, compressed/implicit LUT, spatial-aware, and non-LUT baselines.
Our method, LoR-LUT, is complementary to the above: instead of enlarging or replacing LUTs, we learn compact basis LUTs augmented by \emph{low-rank residual corrections}, preserving standard trilinear/tetrahedral complexity while reducing parameters and improving perceptual fidelity.
\end{itemize}
\section{Conclusion}
\label{sec:conclusion}

We have presented \textbf{LoR-LUT}, a unified low-rank formulation for compact and interpretable 3D lookup table generation.
The motivation of this research comes from Canonical Polyadic Decomposition of tensors in the context of LUT design so that we could address the limitations discussed earlier.  Departing from conventional dense basis fusion, our approach introduces a lightweight \emph{low-rank residual} 
that augments or even replaces base LUTs while maintaining the same trilinear interpolation complexity.
Through this formulation, LoR-LUT achieves an appealing balance between expressiveness, interpretability, and efficiency:
it reproduces expert-level retouching on MIT-Adobe FiveK with sub-megabyte model size and real-time 4K throughput.
Comprehensive experiments demonstrate that the low-rank residual alone captures most of the color transformation manifold,
revealing that photographic adjustments can be effectively modeled within a low-dimensional rank space.
This property not only reduces redundancy in existing adaptive LUT frameworks but also offers a new pathway toward 
transparent and controllable image enhancement.
The accompanying \textbf{LoR-LUT Viewer} further strengthens interpretability by visualizing and manipulating 
rank components interactively.

Looking forward, we plan to extend LoR-LUT to image style transfer and video enhancement domains,
leveraging rank modulation.
We also envision integrating LoR-LUT into hardware-level ISP pipelines and mobile photography systems, 
where its explicit and compact design can provide a practical foundation 
for the next generation of real-time, user-controllable image enhancement models.

\newpage
{
    \small
    \bibliographystyle{ieeenat_fullname}
    \bibliography{main}
}


\end{document}